\documentclass[10pt,twocolumn,letterpaper]{article}

\usepackage{iccv}              
\usepackage{booktabs} 
\usepackage{amsmath}
\usepackage{amsfonts}
\usepackage{comment}
\usepackage{footmisc}
\usepackage{mathrsfs}
\usepackage{graphicx}
\usepackage{multirow}
\usepackage{algorithm}
\usepackage{algorithmicx}
\usepackage{algpseudocode}
\usepackage{multicol}  
\usepackage{enumitem}
\usepackage{array}
\usepackage{float}
\definecolor{iccvblue}{rgb}{0.21,0.49,0.74}
\usepackage[pagebackref,breaklinks,colorlinks,allcolors=iccvblue]{hyperref}
\hypersetup{hidelinks}
\usepackage{bm}
\usepackage{dsfont}
\usepackage{adjustbox}
\usepackage{tcolorbox}

\newcommand{\ours}{\texttt{UV-CoT}}

\newcommand{\tableCellHeight}{1}
\newcommand{\tabstyle}[1]{
  \setlength{\tabcolsep}{#1}
  \renewcommand{\arraystretch}{\tableCellHeight}
  \centering
  \small
}

\def\paperID{XXX} 
\def\confName{ICCV}
\def\confYear{2025}

\title{Unsupervised Visual Chain-of-Thought Reasoning via Preference Optimization}

\author{
  \textbf{Kesen Zhao$^{1}$ \hspace{1em} Beier Zhu$^{1}$\thanks{Corresponding author.  The code is available in \url{https://github.com/kesenzhao/UV-CoT}.} \hspace{1em}
 Qianru Sun$^{2}$ \hspace{1em} Hanwang Zhang$^{1}$} \\
 \small $^1$Nanyang Technological University, $^2$Singapore Management University\\
\texttt{\small kesen002@e.ntu.edu.sg, qianrusun@smu.edu.sg} \\ \texttt{\small \{beier.zhu, hanwangzhang\}@ntu.edu.sg}}

\begin{document}
\maketitle
\begin{abstract}

Chain-of-thought (CoT) reasoning greatly improves the interpretability and problem-solving abilities of multimodal large language models (MLLMs). However, existing approaches focus on text CoT, limiting their ability to leverage visual cues. Visual CoT remains underexplored, and the only work~\cite{shao2025visual} is based on supervised fine-tuning that relies on extensive labeled bounding-box data and is hard to generalize to unseen cases. In this paper, we introduce Unsupervised Visual CoT (\texttt{UV-CoT}), a novel framework for image-level CoT reasoning via preference optimization. \texttt{UV-CoT} performs preference comparisons between model-generated bounding boxes (one is preferred and the other is dis-preferred), eliminating the need for bounding-box annotations. We get such preference data by introducing an automatic data generation pipeline. Given an image, our target MLLM (e.g., LLaVA-1.5-7B) generates seed bounding boxes using a template prompt and then answers the question using each bounded region as input. An evaluator MLLM (e.g., OmniLLM-12B) ranks the responses, and these rankings serve as supervision to train the target MLLM with \texttt{UV-CoT} by minimizing negative log-likelihood losses. By emulating human perception--identifying key regions and reasoning based on them--\texttt{UV-CoT} can improve visual comprehension, particularly in spatial reasoning tasks where textual descriptions alone fall short. Our experiments on six datasets demonstrate the superiority of \texttt{UV-CoT}, compared to the state-of-the-art textual and visual CoT methods. Our zero-shot testing on four unseen datasets shows the strong generalization of \texttt{UV-CoT}. 

\end{abstract}    
\section{Introduction}
\label{sec:intro}

With the recent advancements in multimodal large language models (MLLMs)~\cite{liu2024improved, liu2024visual, bai2023qwen, yu2023reformulating}, many efforts have been made to incorporate text CoT reasoning~\cite{wei2022chain,kojima2022large, zhang2022automatic, feng2024towards} to handle complex vision-language tasks~\cite{zhang2023multimodal, yang2023mm, xu2024llava}. However, the visual understanding ability of MLLM is limited by fixed-granularity image processing, \ie, the MLLM cannot dynamically adjust focus across different spatial regions of the input image, even when guided by text CoT prompts~\cite{huang2024mini}. 
This underscores the critical need to explicitly integrate visual cues into the CoT process.

\begin{figure*}[t]
    \centering
    \includegraphics[width=1\linewidth]{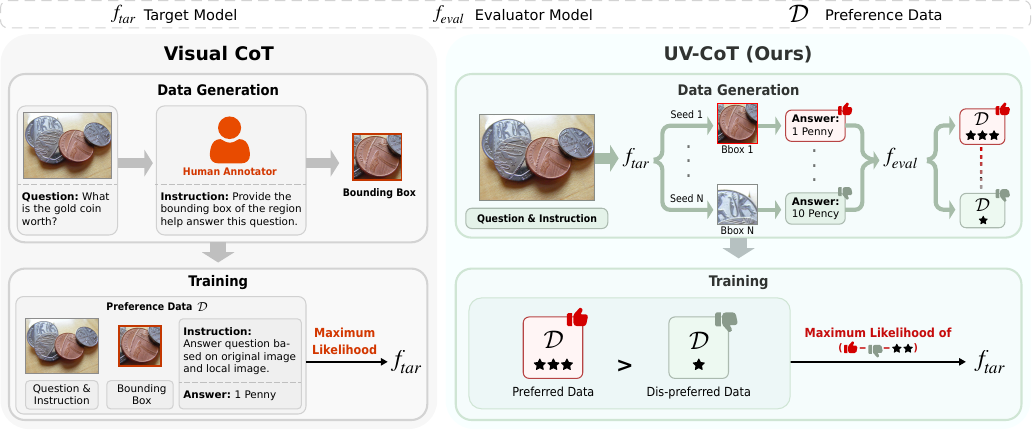}
    \caption{Comparison of Visual-CoT~\cite{shao2025visual} and our \texttt{UV-CoT}. \textbf{Left}: Visual-CoT relies on human-annotated bounding boxes to identify key regions. The model is trained via supervised fine-tuning to maximize the likelihood of the labeled data. \textbf{Right}: \texttt{UV-CoT} eliminates the need for human annotation. Given an image, the target model generates seed bounding boxes and answers questions based on these regions, respectively. An evaluator MLLM then scores the responses as a proxy for assessing region quality. Lastly, the target model is optimized via preference optimization by maximizing the likelihood of preferred regions over dis-preferred ones.}
    \vspace{-2mm}
    \label{fig:overview}
\end{figure*}

A very recent work, Visual-CoT~\cite{shao2025visual}, has made an initial attempt towards this goal.
The model is trained using supervised fine-tuning (SFT) with human-labeled bounding boxes that indicate the key image regions relevant to the question. It performs the multimodal CoT approach with human-annotated reasoning steps by associating textual inputs with the detected regions. An overview of Visual-CoT is presented in~\cref{fig:overview}.
However, it has two key drawbacks: (1) it relies on large-scale, high-quality labeled data, making it costly and hard to scale; and (2) SFT learns only from positive examples (\ie, the labeled data), limiting its ability to generalize to unseen or ambiguous scenarios where intermediate reasoning or dynamic interpretation is needed.

To address these issues, we introduce an \texttt{U}nsupervised approach to \texttt{V}isual \texttt{CoT} dubbed as \texttt{UV-CoT}.
It has two key parts: data collection and model training.
The data collection does not need human annotation, as it leverages the data generation and evaluation capabilities of pre-trained MLLMs.
The model training is inspired by the idea of direct preference optimization (DPO). It is implemented with an adapted version of DPO to address specific limitations in capturing the degree of preference and fine-grained region-based reasoning when conducting visual CoT on MLLMs.
\textbf{Our method} \texttt{UV-CoT}, as shown in~\cref{fig:overview}, differs from Visual-CoT~\cite{shao2025visual} by adopting an unsupervised approach with contrastive preference data. We design an automatic two-step pipeline to generate this data:
1) Region Generation: Given an image, the target model generates multiple seed bounding boxes using a template prompt. Then it answers the question by processing each bounded region together with the question as input.
2) Quality Assessment: An evaluator MLLM scores the responses, using these scores as proxies to measure the quality of the regions.
Unlike traditional DPO, we propose Score-DPO (sDPO), which not only ranks preference data (i.e., preferred and dis-preferred responses shown in \cref{fig:overview}) but also assigns preference scores. This scoring enables more precise optimization based on score differences.
During \texttt{UV-CoT} training, the rankings of the preference data act as supervision by minimizing negative log-likelihood loss, while the scores define the decision margin. By mimicking human perception--first identifying key regions, then reasoning over them--\texttt{UV-CoT} significantly improves visual comprehension, especially in spatial reasoning tasks where text-based methods fall short. By leveraging unsupervised data in a contrastive way, \texttt{UV-CoT} also shows strong generalization ability when tested on unseen datasets. 

\textbf{Our contributions} in this paper include: 1)~an automatic pipeline for generating high-quality preference data, enabling robust and scalable preference learning of \texttt{UV-CoT}; 2)~an improved version of DPO by integrating the degree of preference for visual regions, allowing the model to distinguish key regions more precisely; and 3)~extensive experiments on multiple challenging datasets, demonstrating state-of-the-art performance of \texttt{UV-CoT} on four benchmarks and strong generalization to four unseen test datasets.

\section{Related Works}
\label{sec:relatedwork}

\textbf{Chain-of-thought in LLMs and MLLMs.}
LLMs \cite{lu2024deepseek, touvron2023llama, bai2023qwen,panprecise,hu2024psycollm,jia2024g3}  with CoT \cite{wei2022chain, feng2023towards, chen2025towards} show strong inferential abilities by introducing intermediate reasoning steps. Both manually designed \cite{wei2022chain} and self-generated \cite{kojima2022large, zhang2022automatic} reasoning approaches have proven effective. 
In contrast, MLLMs rely on image encoders \cite{radford2021learning, zhu2024selective,Jiang_2025_CVPR, zhu2024boosting, zhu2024enhancing} to extract visual features but often struggle with reasoning due to inherent differences in how textual and visual data are processed \cite{zhou2025egotextvqa, liu2024multimodal, zhang2024notellm}. Multimodal CoT methods \cite{zhang2023multimodal, yang2023mm, xu2024llava} attempt to address this by transforming multimodal inputs into a unified textual format, enabling LLMs to perform CoT at the text level. However, this transformation introduces significant information loss and prevents the models from capturing key visual details \cite{zhang2023multimodal}. For example, LLaVA-CoT \cite{xu2024llava} leverages GPT-4o to summarize questions and image captions but suffers from weak optical character recognition and sometimes hallucinations.

A very recent work, Visual-CoT~\cite{shao2025visual}, improves the MLLM reasoning by introducing CoT methods at the image level. This approach involves scanning the entire image, identifying key references, and then focusing the model on specific regions for reasoning. Despite its improvements, Visual-CoT is heavily based on costly human-annotated data. In contrast, our \texttt{UV-CoT} framework utilizes unsupervised preference optimization with auto-generated preference data, eliminating the need for manual annotations.

\noindent\textbf{Preference learning in LLMs and MLLMs.}
RLHF \cite{ziegler2019fine} aligns LLMs with human preferences by training a reward model via contrastive response evaluations. To reduce reliance on human annotations, RLAIF \cite{lee2023rlaif} leverages pretrained LLMs for preference label generation. However, RL-based fine-tuning faces stability and efficiency challenges. Direct Preference Optimization (DPO) \cite{rafailov2023direct} addresses this by directly linking reward functions to optimal policies, eliminating reward model fine-tuning. Further improvements include IPO \cite{azar2024general}, which mitigates overfitting with a bounded preference function, and KTO \cite{ethayarajh2024kto}, which removes the need for paired preference data, relying instead on single examples labeled as either `good' or `bad'.

These preference learning techniques are applied to MLLMs, with RLHF-V and RLAIF-V \cite{yu2024rlhf, yu2024rlaif} refining behavior alignments using human and LLM-generated labels. To mitigate reward hacking, \citeauthor{sun2023aligning} \cite{sun2023aligning} enhance reward models with additional factual details, such as image captions and verified choices, further improving the performance of MLLM.
Some works have attempted to apply DPO in the CoT process \cite{pang2025iterative, lai2024step}. However, these methods are designed for only text-level CoT and do not effectively handle visual features or cues.
In contrast, in this paper, we propose \texttt{UV-CoT}, a specialized framework for image-level CoT reasoning inspired by the idea of DPO. 
Unlike traditional DPO, \texttt{UV-CoT} not only ranks preference data (i.e., preferred and dis-preferred data) but also assigns preference scores. This scoring enables more precise optimization of the MLLMs based on score differences.


\section{Method}
\label{sec:method}

\begin{figure}[t]
	\centering
	\includegraphics[width=.95\linewidth]{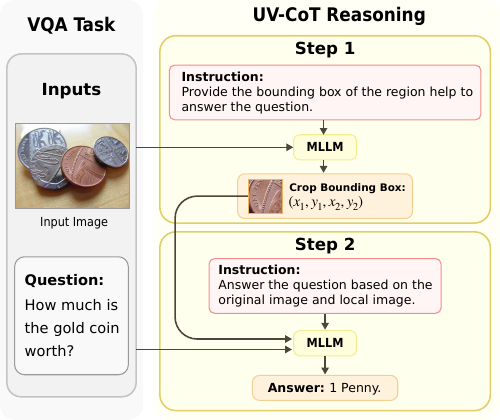}
	\caption{Illustration of \texttt{UV-CoT} reasoning.}
	\label{fig:overview2}
    \vspace{-3mm}
\end{figure}

 

\cref{fig:overview2} illustrates the pipeline of \texttt{UV-CoT} reasoning. 
Given the original image and the question, we append a CoT prompt to guide the target MLLM in identifying the most informative image region and specifying its location via bounding box coordinates. A visual sampler then extracts the bounded region from the image. The MLLM subsequently integrates visual tokens from original and cropped images to generate more precise and comprehensive answers. 
In~\cref{sec:data}, we detail the automatic preference data generation pipeline.
In~\cref{sec:learning}, we describe our Score-DPO (sDPO) for image-level CoT reasoning.


\begin{algorithm}[t]
\caption{\label{alg:process} Preference data generation for a query $x$}
\begin{algorithmic}[1]
    \State \textbf{Input:} Target model $f_\mathsf{tar}$, evaluator $f_\mathsf{eval}$, an image-question query  $x$, number of seeds $n$, and number of preference pairs $k$.
    \State \textbf{Output:} Preference data $\mathcal{D}$
    \State Initialize  $y_0=x$
    \For{$t = 1$ to $T$}
        \State $\{y_t^i\}_{i=1}^n \gets \mathsf{Generate}(f_\mathsf{tar}, y_{0:t-1}, n)$
        \State $\{s^i\}_{i=1}^n \gets \mathsf{Evaluate}(f_\mathsf{eval}, y_{0:t-1}, \{y_t^i\}_{i=1}^n)$
        \State $\mathcal{D}_t \gets \mathsf{ConstructPairs}(y_{0:t-1},\{y_t^i\}_{i=1}^n,\{s^i\}_{i=1}^n, k)$
        \State $y_t \gets \mathsf{Select}(y_{0:t-1},\{y_t^i\}_{i=1}^n,\{s^i\}_{i=1}^n)$    
    \EndFor

    \State \Return $\mathcal{D}=\{\mathcal{D}_1,\dots,\mathcal{D}_T\}$
\end{algorithmic}
\end{algorithm}

\subsection{Preference Data Generation} \label{sec:data}
Given a target model $f_\mathsf{tar}$, an evaluator model $f_\mathsf{eval}$ (the target MLLM can also serve as the evaluator, validated in \cref{table:evaluator}), and an image-question pair $x$, we illustrate how to construct $n$ preference data points. 
Assuming there are $T$ steps in CoT reasoning process, we generate preference data for $T$ times on the way, as described in~\cref{alg:process}.
At each timestep $t$ (\ie, a reasoning step $t$), the process includes four stages: Response Generation, Response Evaluation, Pair Construction, and Response Selection. 

\noindent\textbf{Response generation.} 
%
The goal of this stage is to generate seed bounding boxes and produce intermediate responses to the question using the target model. Here, a `response' refers to any model output at an intermediate step, not necessarily the final answer to the question. We denote the model’s response at timestep $t$ as $y_t$, with the initial input $x$ represented as $y_0$. To encourage diversity in the bounding boxes and subsequent responses, we apply stochastic decoding to the target model $f_\mathsf{tar}$ with $n$ different random seeds, resulting in a set of responses $\{y_t^i\}_{i=1}^n$.

\noindent\textbf{Response evaluation.} 
This stage evaluates the quality of all generated responses. The evaluator model not only scores each individual response but also counts how this response influences the quality of its next response in the chain. This cumulative evaluation approach helps quantify the impact of each bounded region on the overall reasoning process.
%
%
Below, we elaborate on the formulation.
At timestep $t$, the evaluator assigns scores for $y_t^i$ as follows:
\begin{equation} 
\begin{aligned} 
s_{\mathsf{cur}}^i &=f_\mathsf{eval}(y^i_t \mid y_{0:t-1}), \\ 
s_{\mathsf{nxt}}^i &=\mathbb{E}[f_\mathsf{eval}(y_{t+1}^1 \mid y_{0:t-1}, y^i_t)], \\ 
s^i &=s_{\mathsf{cur}}^i+\gamma s_{\mathsf{nxt}}^i,
\end{aligned}
\end{equation} 
where $s_{\mathsf{nxt}}^i$ reflects the impact on the next response and $\gamma>0$ is a hyperparameter to combine the current and next response scores, with $\gamma=0$ at the last step. We estimate the expectation $\mathbb{E}[\cdot]$ by randomly sampling next responses.

\noindent\textbf{Pair construction.} At each timestep $t$, we randomly select $k$ (preferred and dis-preferred) pairs from  $\{y_t^i\}_{i=1}^n$. 
For a single pair, we concatenate the preferred response with the past response chain $y_{0:t-1}$ (preserved after $t\!-\!1$ timesteps) and get a ``preferred chain" denoted as $y_t^w$, and then we concatenate the dis-preferred response in the same way to get a `dis-preferred chain' denoted as $y_t^l$.
The pair of chains also includes their respective scores, and they are represented as $\{y_w, s_w, y_l, s_l\}$.
The overall $k$ pairs of chains compose the preference dataset $\mathcal{D}_t$.

\noindent\textbf{Response selection.}
The abovementioned `past response chain $y_{0:t-1}$' is unique and is concatenated by the highest-scoring response at timestep $t\!-\!1$ and the preserved chain at timestep $t\!=\!2$, \ie, $y_{0:t-2}$. In other words, when finishing each timestep process, we preserve only the best chain and use it for the next step.
%

\subsection{Unsupervised Learning of \texttt{UV-CoT}}\label{sec:learning}
Assume the preference dataset $\mathcal{D}$ has been generated across $t$ timesteps, we then optimize the target model with our \texttt{UV-CoT} via preference optimization on $\mathcal{D}$.
DPO~\cite{amini2024direct} is widely used in preference learning, and it ranks responses without quantifying preference intensity. 
In our case of image-level reasoning, key regions vary in influence, necessitating finer differentiation between responses.
Therefore, we refine DPO by adjusting the margin to capture the key region's influence. 	We name our loss Score-DPO, abbreviated as \textbf{sDPO}, as it incorporates the preference score into the optimization. The loss is formulated as:

\begin{equation} \label{eq:dpo_ours}
\begin{aligned}
&\mathcal{L}_\text{sDPO}(\theta) = -\underset{\left(x, y_w, y_l\right) \sim \mathcal{D}}{\mathbb{E}} \left[ \log \sigma \left( \beta \log \frac{\pi_{\theta}(y_w \mid x)}{\pi_{\mathrm{ref}}(y_w \mid x)} \right. \right. \\
&\quad \left. \left. - \beta \log \frac{\pi_{\theta}(y_l \mid x)}{\pi_{\mathrm{ref}}(y_l \mid x)} - (g(s_w) -g(s_l)) \right) \right],
\end{aligned}
\end{equation}
where  $\pi_\theta $ is the target model, and  $\pi_\text{ref}$  is its frozen initialization, serving as a reference to constrain deviation from the original model.
 $g(\cdot):\mathbb{R} \rightarrow \mathbb{R}$ is a monotonically increasing function that maps preference scores into the logit space of the DPO objective. 
 
 To provide a deeper understanding of our sDPO loss, we establish its connection to the standard DPO loss.
DPO reformulates reward model training as policy optimization by reparameterizing the reward function in PPO \cite{schulman2017proximal}:
\begin{equation} 
r(x, y)=\beta \log \frac{\pi_{\theta}(y \mid x)}{\pi_{\mathrm{ref}}(y \mid x)}+\beta \log Z(x)
\end{equation} 
Using the Bradley-Terry~\cite{bradley1952rank} model, the standard DPO aims to minimize the negative log-likelihood of the difference of rewards between paired responses:
\begin{equation}
\begin{aligned}
& {P}\left(y_w-y_l>0\right)=\sigma\left(r\left(x, y_w\right)-r\left(x, y_l\right)\right) \\
& \mathcal{L}_\text{DPO}=-\mathbb{E}_{(x,y_w,y_l)\sim \mathcal{D}}\left[\log {P}\left(y_w-y_l>0\right) \right],
\end{aligned}
\end{equation} 
where $\sigma(x)=\frac{1}{1+\exp(-x)}$ is the sigmoid function. 

Let $\Delta_{r}=g(s_w)-g(s_l)$ and define the Gumbel-distributed random variables $R_w \sim \operatorname{Gumbel}\left(r\left(x, y_w\right), 1\right)$ and $R_l \sim \operatorname{Gumbel} \left(r\left(x, y_l\right), 1\right)$. Then, we derive:
\begin{equation} 
\begin{aligned}
&P\left(R_w-R_l>\Delta_{r}\right) =\sigma\left(r\left(x, y_w\right)-r\left(x, y_l\right) - \Delta{r}\right) \\
& =\sigma\left(\beta \log \frac{\pi^*\left(y_w \mid x\right)}{\pi_{\mathrm{ref}}\left(y_w \mid x\right)}-\beta \log \frac{\pi^*\left(y_l \mid x\right)}{\pi_{\mathrm{ref}}\left(y_l \mid x\right)}- \Delta{r}\right)
\end{aligned}
\end{equation} 
This result follows from the definition of Gumbel random variables \cite{amini2024direct} and the Gumbel-max trick \cite{Maddison2017Gumbel}. 
We provide a detailed proof in Appendix B.
By maximizing the log-likelihood, we obtain our proposed loss function.
The Gumbel distribution models the extreme values of a variable, while $\Delta_{r}$ quantifies the degree of difference between preference pairs. Thus, our loss function explicitly optimizes preference learning by distinguishing not only the order but also the magnitude of preference differences. 

\begin{algorithm}[t]
\caption{\label{alg:iterative} Iterative learning of \texttt{UV-CoT}}
\begin{algorithmic}[1]
    \State \textbf{Input:} Initial target model $f^1_\mathsf{tar}$, evaluator model $f_\mathsf{eval}$, $\mathcal{X} = \{\mathcal{X}_1, \dots, \mathcal{X}_m\}$, each $\mathcal{X}_i$ is a subset of image-question query 
    \State \textbf{Output:} $f^m_\mathsf{tar}$
    
    \For{$i = 1$ to $m$}
        \State $\mathcal{D}_i \gets \mathsf{GenerateData}(f^i_\mathsf{tar}, f_\mathsf{eval}, \mathcal{X}_i)$ \Comment{\cref{alg:process}}
        \State $f^{i+1}_\mathsf{tar} \gets \mathsf{Train}(f^i_\mathsf{tar}, \mathcal{D}_i)$   \Comment{~\cref{eq:dpo_ours}} 
    \EndFor

    \State \Return $f^m_\mathsf{tar}$
\end{algorithmic}
\end{algorithm}

\noindent\textbf{Iterative learning.} Standard DPO relies on static preference data during training, which can lead to distributional mismatch between training data and the model's generated outputs. To mitigate this issue, we adopt an iterative learning approach inspired by \cite{yu2024rlaif}. 
\cref{alg:iterative} outlines the iterative learning process of \texttt{UV-CoT}, which incrementally refines a target model through preference learning. The iteration repeats  $m$  times, and the total image-question query set  $\mathcal{X}$ is evenly divided into  $m$  subsets,  $\mathcal{X} = \{\mathcal{X}_1, \ldots, \mathcal{X}_m\}$ , each assigned to one iteration.
 The process starts with an initial target model $f_\mathsf{tar}$, an evaluator model $f_\mathsf{eval}$, and $\mathcal{X}$.
In each iteration $i$, the algorithm first generates preference data $\mathcal{D}_i$ using the current target model $f^i_\mathsf{tar}$ and the subset $\mathcal{X}_i$, using the procedure in \cref{alg:process}. This newly generated preference data is then used to train the next iteration of the target model $f^{i+1}_\mathsf{tar}$ using our \texttt{UV-CoT} loss of~\cref{eq:dpo_ours}. The process continues until the final model $f^m_\mathsf{tar}$ is obtained.
By dynamically updating the preference data, our approach ensures that the learned model adapts to its evolving distribution, enhancing training robustness.

\begin{table*}[t] 
\center
\tabstyle{9pt}
\begin{tabular}{c|cccccc|c}
\toprule[1pt]
MLLM          & DocVQA & TextVQA & InfographicsVQA & Flickr30k & GQA & VSR & Average\\ \midrule
LLaVA-1.5-7B  & 0.198 & 0.507 & 0.131 & 0.539 & 0.480 & 0.504 & 0.393\\
LLaVA-1.5-13B & 0.225 & 0.543  & 0.169 & 0.607 & 0.506 & 0.512 & 0.427\\
MiniCPM-o-8B  & 0.232 & 0.529 & 0.175 & 0.618 & 0.495 & 0.521 & 0.428\\ \midrule
OmniLMM-12B  & 0.254 & 0.578 & 0.172 & 0.621 & 0.509 & 0.523 & 0.443\\ \midrule
Visual-CoT-7B (100\% label) & \textbf{0.294} & 0.673 & \underline{0.194} & \textbf{0.652} & \underline{0.546} & 0.532 & \underline{0.482}\\ \midrule
\texttt{UV-CoT} (0\% label) & 0.265 & \underline{0.686} & 0.173 & 0.632 & 0.536 & \underline{0.548} & 0.473\\ 
\texttt{UV-CoT} (10\% label) & \underline{0.283} & \textbf{0.711} & \textbf{0.198} & \underline{0.649} & \textbf{0.568} & \textbf{0.553} & \textbf{0.494}\\ \bottomrule[1pt]
\end{tabular}
\caption{\textbf{Overall comparison} of different models on six evaluation benchmarks. The \textbf{best} result is bold, the \underline{second-best} is underlined. `\%' indicates the percentage of supervised data used in {Visual-CoT}. 	By default, our \texttt{UV-CoT} uses only unsupervised data.}
	\label{table:overall}
\end{table*}
\section{Experiments}
\label{sec:experiments}

\begin{table}[t] \center
\tabstyle{5pt}
\begin{tabular}{c|ccc|c}
\toprule[1pt]
MLLM          & DUDE & SROIE & Visual7w & Average\\ \midrule
LLaVA-1.5-7B  & 0.165 & 0.147 &  0.340 & 0.217 \\ 
LLaVA-1.5-13B & 0.174 & 0.159 & 0.352 & 0.228\\
MiniCPM-o-8B  & 0.182 & 0.165 & 0.341 & 0.229\\ 
OmniLMM-12B  & 0.194 & 0.166 & 0.357 & 0.239\\
Visual-CoT-7B & 0.206 & 0.181 &  0.397& 0.261 \\ \midrule
\texttt{UV-CoT} & \underline{0.241} & \underline{0.184} & \underline{0.432} & \underline{0.286}\\ 
\texttt{UV-CoT}$^*$ &  \textbf{0.253} &  \textbf{0.227}   & \textbf{0.455} & \textbf{0.312}\\
\bottomrule[1pt]
\end{tabular}
\caption{\textbf{Zero-shot experiments} on DUDE, SROIE and Visual7w. 
`\texttt{UV-CoT}$^*$' denotes our model trained with additional unlabeled preference data from the three zero-shot datasets.}
	\label{table:zero}
\end{table}

\subsection{Setup}
\label{sec:setup}
\textbf{Datasets.}
For a comprehensive evaluation, we adopt ten datasets spanning five domains:
{(1) Text/Document}: DocVQA \cite{mathew2021docvqa}, TextVQA \cite{singh2019towards}, DUDE \cite{van2023document}, and SROIE \cite{huang2019icdar2019}.
    {(2) Chart}: InfographicsVQA \cite{mathew2022infographicvqa}. 
    {(3) General VQA}: Flickr30k \cite{plummer2015flickr30k} and Visual7W \cite{zhu2016visual7w}.
    {(4) Relation Reasoning}: VSR \cite{liu2023visual} and GQA \cite{hudson2019gqa}.
    {(5) High-Resolution Image Reasoning}: V$^*$ Bench \cite{wu2024v}.
Notably, we also provide a model trained on data excluding DUDE, SROIE, Visual7W, and V$^*$ Bench, which is used to evaluate zero-shot performance.
See Appendix C.1 for details.

\noindent\textbf{Evaluation.} 
Following prior work~\cite{liu2024visual}, we prompt GPT-4o~\cite{openai2023chatgpt} to assign a score between 0 and 1, with higher scores indicating better prediction. See details in Appendix C.2.

\noindent\textbf{Baselines.} We compare \ours~with five baselines. LLaVA-1.5-\{7B, 13B\}~\cite{liu2024improved} and OmniLMM-12B are strong general baselines. MiniCPM-o-8B \cite{yao2024minicpm} adopts adaptive visual encoding for fine-grained understanding and Visual-CoT-7B \cite{shao2025visual} learns image-level CoT via SFT.

\noindent\textbf{Implementation details.} 
We use LLaVA-1.5-7B as the target model and OmniLMM-12B as the evaluator. To ensure scalability, we avoid using GPT-4 as the evaluator, preventing constraints imposed by high API costs.
We implement iterative learning of \texttt{UV-CoT} over four iterations, utilizing a total of 249K preference data pairs. Notably, \texttt{UV-CoT} achieves higher data efficiency than Visual-CoT, which uses 376K data pairs.
For each iteration, we train the model with AdamW optimizer for 4 epochs with a learning rate of $5 \times 10^{-7}, \beta = 0.1$, and a batch size of 8. In total, data generation takes 80 hours and training requires 60 hours, both conducted on an 8$\times$A100 40GB machine.
Additionally, we provide a variant \texttt{UV-CoT} trained with extra SFT on 10\% of the labeled Visual-CoT data, denoted as \texttt{UV-CoT} (10\%).

\begin{table}[t] \center
\tabstyle{4pt}
\begin{tabular}{c|ccc|c}
\toprule[1pt]
MLLM          & Attributes & GPT4V-hard & OCR & Average\\ \midrule
LLaVA-1.5-7B  & 0.317 & 0 &  0.1 & 0.139 \\ 
LLaVA-1.5-13B & 0.326 & 0.118 & 0.133 & 0.192\\
MiniCPM-o-8B  & 0.322 & 0.118 & 0.1 & 0.180\\ 
OmniLMM-12B  & 0.326 & 0.118 & 0.167 & 0.204\\
Visual-CoT-7B & 0.330 & 0.118 &  0.593& 0.347 \\ \midrule
\texttt{UV-CoT} & \textbf{0.352} & \textbf{0.176} & \textbf{0.677} & \textbf{0.402}\\ 
\bottomrule[1pt]
\end{tabular}
\caption{\textbf{Zero-shot experiments} on V$^*$ Bench (high-resolution image reasoning task, average resolution $2246\times 1582$). }
	\label{table:resolution}
\end{table}

\begin{table*}[t] 
\center
\tabstyle{9pt}
\begin{tabular}{l|cccccc|c}
\toprule[1pt]
Model         & DocVQA & TextVQA & InfographicsVQA & Flickr30k & GQA & VSR & Average \\ \midrule
\texttt{UV-CoT} (10\% labels)  & 0.283 & 0.711 & 0.198 & 0.649 & 0.568 & 0.553 & 0.494\\ \midrule
\texttt{IF:} w/o \texttt{UV-CoT} & 0.149 & 0.574 & 0.160 & 0.585 & 0.509 & 0.522 & 0.417\\
\texttt{IF:} \texttt{UV-CoT} w/ G.T. BBox  & 0.528 & 0.769 & 0.504 & 0.655 & 0.664 & 0.585 & 0.618 \\ 
\texttt{Tr:} w/ naive DPO & 0.258 & 0.695 & 0.189 & 0.623 & 0.552 & 0.534 & 0.475\\ 
\texttt{Tr:} w/o iterative learning & 0.256 & 0.671 & 0.162 & 0.609 & 0.523 & 0.531 & 0.459\\ 
\texttt{Ge:} w/o $\gamma$ & 0.247 & 0.539 & 0.152 & 0.551 & 0.484 & 0.460 & 0.406\\ \bottomrule[1pt]
\end{tabular}
\caption{
\textbf{Ablation study} on key components of \texttt{UV-CoT} (10\% labeled data). `w/o \texttt{UV-CoT}' denotes standard inference without CoT reasoning. `\texttt{UV-CoT} w/ G.T. BBox' uses annotated ground truth bounding boxes. `w/ naive DPO' applies the standard DPO loss. `w/o iterative learning' generates preference pairs for the entire set of question queries $\mathcal{X}$ in a single pass and trains once. `w/o  $\gamma$' evaluates responses with $\gamma = 0$. \texttt{IF}, \texttt{Tr}, and \texttt{Ge} indicate ablations on the inference process, training loss, and data generation, respectively.}
\label{table:ablation}
\end{table*}

\subsection{Comparison with State-of-the-Art Methods}
\label{sec:main_results}

The overall performance comparisons are reported in \cref{table:overall}, leading to the following key observations:

\noindent \textbf{Explicitly incorporating visual cues proves beneficial for multimodal reasoning.} For instance, MiniCPM-o-8B, which uses rule-based cropping to focus on salient regions, outperforms LLaVA-1.5-7B by an average of 3.5\%. However, its reliance on heuristics limits adaptability.
In contrast, image-level CoT models like Visual-CoT and our \texttt{UV-CoT} achieve  greater performance gains, even surpassing the larger LLaVA-1.5-13B, by leveraging MLLMs to adaptively generate key visual regions. This highlights the superior effectiveness of image-level CoT reasoning.

\noindent\textbf{\ours~fundamentally differs from distillation/pseudo-labeling.}
Unlike distillation, where student performance is typically bounded by the teacher, \ours~outperforms its evaluator OmniLMM-12B by 5.1\% on average.
This suggests that \ours~goes beyond simply mimicking a larger model. Direct generation of accurate bounding boxes remains a challenge for MLLMs due to the need for precise spatial localization. Instead, \ours~reformulates the task as  ranking—an inherently simpler and more tractable problem—which leads to better performance.

\noindent\textbf{\ours~outperforms the supervised Visual-CoT.} Despite using significantly less data (249K unlabeled vs. 376K labeled), \texttt{UV-CoT} outperforms Visual-CoT-7B on TextVQA (+1.3\%) and VSR (+1.6\%).
Moreover, \texttt{UV-CoT}(10\%)  surpasses Visual-CoT-7B by 2.1\% on average, with notable gains on TextVQA (+2.5\%), GQA (+2.2\%), and VSR (+2.1\%) and achieves comparable or better performance on the remaining datasets. This validates the effectiveness of our high-quality preference data generation and enhanced preference optimization method. 

\subsection{Zero-Shot Generalization}
\label{sec:zero_shot}

We evaluate the zero-shot performance of \ours~on the test sets of SROIE, DUDE, Visual7W, and V$^*$ Bench \cite{wu2024v}, \textit{without any training exposure to these datasets}.
V$^*$ Bench is a high-resolution benchmark (avg. size: $2246\times 1582$) covering diverse tasks; we focus on three representative ones: Attributes (object attribute recognition), GPT4V-Hard (complex visual reasoning), and OCR.
Additionally, we train a variant of our model, \texttt{UV-CoT}$^*$, using preference data generated from the training splits of SROIE, DUDE, and Visual7W. The results are reported in \cref{table:zero} and \cref{table:resolution}, leading to the following observations:

\noindent\textbf{\ours~exhibits stronger zero-shot performance.}
Supervised CoT learning relies on labeled data and often overfits to specific annotation distributions, limiting its generalization to unseen tasks.
In contrast, \texttt{UV-CoT} uses preference optimization based on relative comparisons, avoiding reliance on absolute labels and enhancing generalization.
As a result, \texttt{UV-CoT} outperforms all baselines across zero-shot datasets (+2.5\% on average), with notable gains on DUDE and Visual7W (both +3.5\%).
 Furthermore, \texttt{UV-CoT}$^*$ achieves greater gains (5.1\% on average), showing that our model can effectively learn the image-level CoT process without the need for costly human annotation.

\noindent\textbf{\ours~excels in high-resolution image reasoning.} Image-level CoT methods—\texttt{UV-CoT} (+19.8\%) and Visual-CoT-7B (+14.3\%)—significantly outperform non-CoT baselines, with over 50\% gains on OCR tasks.
\texttt{UV-CoT} further surpasses Visual-CoT-7B by 5.5\% on average, achieving the best performance across all tasks.
This large performance gap underscores the advantage of unsupervised image-level CoT for high-resolution visual reasoning.

\begin{figure*}[t]
	\centering
	\includegraphics[width=.95\linewidth]{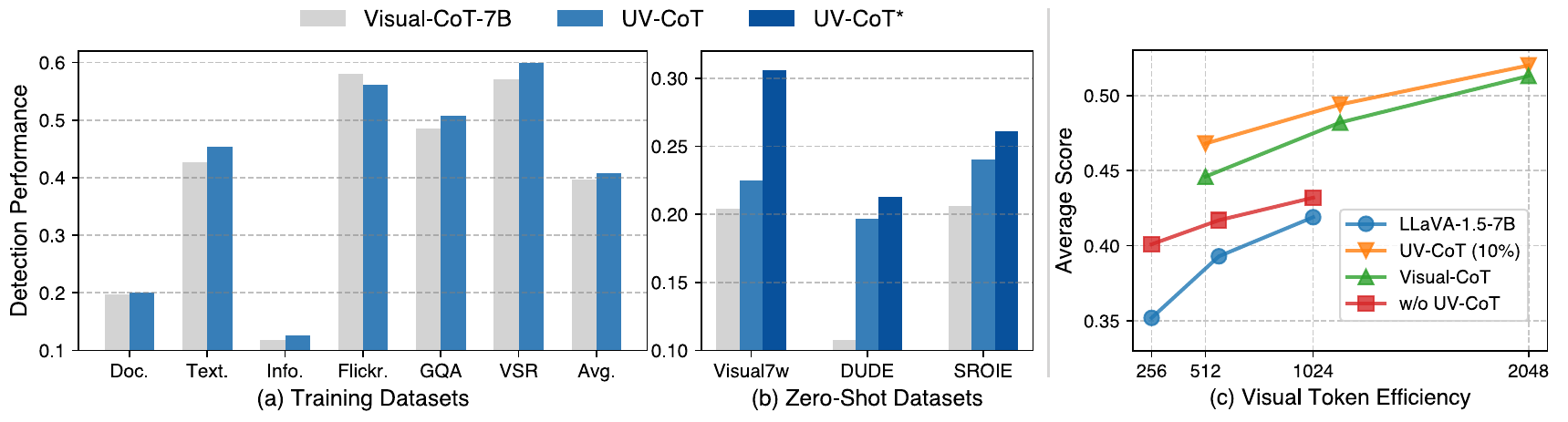}
	\caption{\textbf{(a\&b) Bounding box evaluation} on (a) training datasets and (b) zero-shot datasets. Our \ours~performs better than Visual-CoT. \textbf{(c) Model performance} under varying visual token sizes. Our \ours~demonstrates better token efficiency.}
	\label{fig:detection}
\end{figure*}

\begin{table*}[t] \center
\tabstyle{5pt}
\begin{tabular}{l|cccccc|c}
\toprule[1pt]
Model         & DocVQA & TextVQA & InfographicsVQA & Flickr30k & GQA & VSR & Average \\ \midrule
OmniLMM-12B  & 0.254 & 0.578 & 0.172 & 0.621 & 0.509 & 0.523 & 0.443\\ 
\texttt{IF:} OmniLMM-12B + CoT & 0.305 & 0.722 & 0.217 & 0.675 & 0.580 & 0.595 & 0.516\\ \midrule
LLaVA-1.5-7B & 0.198 & 0.507 & 0.131 & 0.539 & 0.480 & 0.504 & 0.393 \\ 
\texttt{IF:} LLaVA-1.5-7B + CoT & 0.245 & 0.577 & 0.149 & 0.606 & 0.530 & 0.533 & 0.440\\ \midrule
\texttt{Tr:} \texttt{UV-CoT} (Evaluator: self-evaluated)  & 0.242 & 0.609 & 0.153 & 0.598 & 0.517 & 0.526 & 0.441\\ 
\texttt{Tr:} \texttt{UV-CoT} (Evaluator: OmniLMM-12B) & 0.265 & 0.686 & 0.173 & 0.632 & 0.536 & 0.548 & 0.473\\ \bottomrule[1pt]
\end{tabular}
\caption{
Analysis of evaluator model. `+ CoT' refers to inference assisted by bounding boxes generated by \texttt{UV-CoT}. Self-evaluated denotes using target model as the evaluator during training. \texttt{IF} and \texttt{Tr} indicate experiments conducted on the inference and training process.}
\label{table:evaluator}
\end{table*}

\subsection{Ablation Studies}
\label{sec:ablation}
In \cref{table:ablation}, we present the ablations on key components.

\noindent\textbf{Image-level CoT:} We design two variants to evaluate the impact of image-level CoT:
(1) `w/o \ours' removes intermediate reasoning and directly outputs answers;
(2) `\ours~w/ G.T. BBox' replaces predicted regions with ground truth bounding boxes to assess localization quality.
Results show that removing CoT leads to a significant drop (–7.7\% on average), confirming its necessity.
On Flickr30k, \ours~matches the G.T. variant, suggesting accurate region selection.
However, in DocVQA and InfographicsVQA, using G.T. boxes yields better performance, revealing the difficulty of precise localization.
This highlights the potential of our method and suggests future work for improving region accuracy for complex tasks.

\noindent \textbf{Score-DPO (sDPO):} `w/ naive DPO' applies standard DPO loss and shows degraded performance  across all datasets (1.9\% on average), especially on DocVQA (2.5\%) and Flickr30k (2.6\%), revealing its limitations for CoT learning. In contrast, our sDPO incorporates preference scores to better differentiate choices, yielding consistent gains.

\noindent\textbf{Iterative learning:} `w/o iterative learning' generates all preference pairs in a single pass and trains only once, leading to a significant performance drop (3.5\% on average). 
This underscores the importance of iterative learning in continuously aligning the preference data distribution with the model's evolving policy throughout the training process.

\noindent \textbf{Response evaluation:}  `w/o $\gamma$ model' sets  $\gamma=0$ during response evaluation stage, ignoring  the next response when generating preference scores. 
We observe a significant performance drop (8.8\% on average), particularly on TextVQA (17.2\%), highlighting the difficulty MLLMs face in directly evaluating bounding boxes. This  underscores the necessity of incorporating next response in our evaluation method.

\begin{figure*}[t]
	\centering
	\includegraphics[width=.9\linewidth]{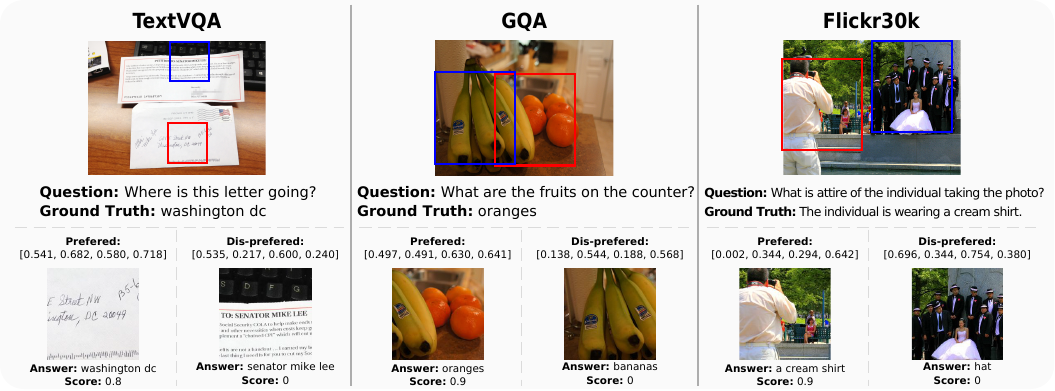}
	\caption{Visualization of preference data generated by~\cref{alg:process}. Preferred BBoxes are in {\color{red} \textbf{red}}. Dis-preferred BBoxes are in {\color{blue} \textbf{blue}}.}
	\label{fig:case1}
\end{figure*}


\begin{figure*}[t]
	\centering
	\includegraphics[width=.9\linewidth]{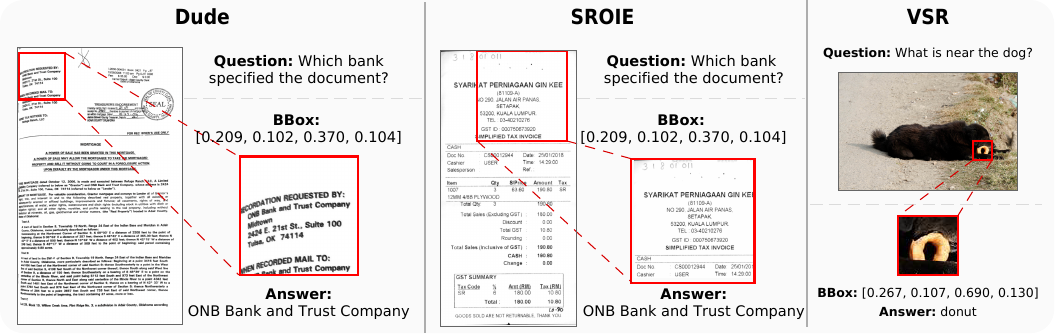}
	\caption{Visualization of our $\texttt{UV-CoT}$ inference. Model-generated bounding boxes are shown in {\color{red} \textbf{red}}.}
	\label{fig:case2}
    \vspace{-3mm}
\end{figure*}

\subsection{Bounding Box Evaluation}
\label{sec:detection}
We compare the quality of bounding boxes learned from \textit{supervised} and \textit{unsupervised}  strategies using GPT-4o as a scorer, on both training datasets (\cref{fig:detection}~(a)) and zero-shot datasets (\cref{fig:detection}~(b)).
Our main observations are:

\noindent \textbf{(1)} Our \texttt{UV-CoT} outperforms Visual-CoT-7B, achieving higher scores in five of six datasets, supporting its superior performance in generating helpful bounding box.

\noindent \textbf{(2)} The bounding box quality is closely related to the final performance. Our model exhibits lower scores (below 0.210) for bounding box generation in DocVQA and InfographicsVQA, correlating with its reduced final scores in these datasets (below 0.290 in \cref{table:overall}). It underscores the validity of evaluating bounding box quality through its impact on subsequent answers. 

\noindent \textbf{(3)} Both \texttt{UV-CoT} and \texttt{UV-CoT}$^*$ outperform Visual-CoT-7B across all zero-shot datasets, which illustrates the strong generalization of our method in bounding box generation.

\subsection{Insight of Evaluator Model}
\label{sec:evaluator}

We further perform studies to better understand the role of evaluator model. Key findings from \cref{table:evaluator} are:

\noindent \textbf{(1)} We compare \texttt{UV-CoT} with its self-evaluated variant, where the evaluator model is the same as the target model (initialed with LLaVA-1.5-7B). Although the self-evaluated version exhibits a performance decrease of 3.2\% compared to the original \texttt{UV-CoT}, it still outperforms LLaVA-1.5-7B (+4.8\% on average) across all evaluated datasets and achieves performance comparable to the larger OmniLMM-12B model (-0.2\% on average). This demonstrates that \texttt{UV-CoT} maintains robust performance even under self-evaluation, highlighting its efficiency without requiring larger model scales.

\noindent \textbf{(2)} For OmniLMM-12B and LLaVA-1.5-7B, we incorporate a CoT process using bounding boxes generated by \texttt{UV-CoT}. The CoT-enhanced versions significantly outperform their original counterparts, achieving average performance gains of 4.7\% for LLaVA-1.5-7B and 7.4\% for OmniLMM-12B. Remarkably, these models were not fine-tuned for the CoT process, indicating that the bounding box information alone substantially improves performance. This finding underscores that our evaluating process simplifies the task of generating complex spatial annotations, enabling MLLMs to focus on evaluating final answers.

\subsection{Other Detailed Analyses}
\label{sec:analysis}

\noindent\textbf{Visual token efficiency.}
Compared to standard MLLM generation, image-level CoT doubles the number of visual tokens by processing additional local image regions. To evaluate token efficiency, \ie, performance under the same visual token budget, we resize the input image to different resolutions ($224^2, 336^2$ and $448^2$) and report the average score of different models in \cref{fig:detection}(c). Our key findings are:

\noindent\textbf{(1)} MLLMs with image-level reasoning (Visual-CoT-7B and our \texttt{UV-CoT}) demonstrate better token efficiency than the standard answer generation pipeline. \textit{E.g.}, they achieve higher performance with 512 visual tokens than the standard pipeline does with 1024 tokens.

\noindent\textbf{(2)} Our \texttt{UV-CoT} consistently outperforms Visual-CoT-7B across all scales, achieving higher average scores. This highlights the token efficiency of our method.




\noindent\textbf{Visualization.}
\cref{fig:case1} visualizes some preference data from our \texttt{UV-CoT} inference process. Given different local regions and their corresponding answers, the evaluator MLLM assigns reasonable scores, validating the effectiveness of our automatic generation and labeling process. 
In \cref{fig:case2}, we present reasoning cases with model-generated bounding boxes overlaid. The precision of bounding box detection and the depth of understanding play a crucial role in determining the quality of generated answers.


\section{Conclusion}
\label{sec:conclusion}
In this work, we propose \texttt{UV-CoT}, a framework that enables image-level CoT reasoning in MLLMs via preference optimization. Unlike previous methods that rely on SFT needing large amounts of labeled data, our approach leverages  unsupervised learning to refine the model's ability with image-level CoT using model-generated preference data (which are rough but useful). We address key challenges in preference data generation and effective optimization, ensuring a more adaptive and interpretable reasoning process. Extensive experiments demonstrate that \texttt{UV-CoT} achieves state-of-the-art performance, significantly improving visual comprehension in MLLMs on ten reasoning datasets. 
Our findings highlight the potential of preference learning as a scalable alternative to traditional SFT, enabling more robust and data-efficient multimodal reasoning.

\section*{Acknowledgements}
This research is supported by the National Research Foundation, Singapore under its AI Singapore Programme (AISG Award No: AISG3-RP-2022-030).
Thanks to Zichen Tian for his assistance with figure visualization throughout this work. The authors also thank the reviewers for their valuable comments and suggestions.


{
    \small
    \bibliographystyle{ieeenat_fullname}
    \bibliography{main}
}
\clearpage
\appendix
\section{Outline}
We begin by presenting an overview of our Appendix.
\begin{itemize}[leftmargin=*]
\item \textbf{Section B: Framework Datails.} We provide a detailed explanation on the connection between our UV-CoT and DPO.
\item \textbf{Section C: Implement Details.} We describe the specifics of our dataset, and evaluation methodology.
\item \textbf{Section D: Limitations.} We analysis the constraints and challenges of our approach.
\item \textbf{Section E: Potential negative societal impacts.} We discuss possible negative consequences and ethical considerations associated with our work.

\end{itemize}

\section{Connection between UV-CoT and DPO}
\subsection{Loss Function Formulation}
To better captures the impact of key regions, we introduce a preference-weighted optimization approach. 
The loss function for UV-CoT is defined as follows:
\begin{equation} \label{eq:uvcot_loss}
\begin{aligned}
\mathcal{L}_{{sDPO}}(\theta) &= -\mathbb{E}_{(x, y_w, y_l) \sim \mathcal{D}}  
\Bigg[ \log \sigma \Bigg( \beta \log \frac{\pi_{\theta}(y_w \mid x)}{\pi_{\mathrm{ref}}(y_w \mid x)}  \\
&\quad - \beta \log \frac{\pi_{\theta}(y_l \mid x)}{\pi_{\mathrm{ref}}(y_l \mid x)} - \big( g(s_w) - g(s_l) \big) \Bigg) \Bigg].
\end{aligned}
\end{equation}

where \(\pi_{\theta}\) is the target policy model being optimized, \(\pi_{\mathrm{ref}}\) is the frozen reference model, constraining deviations from the initial policy,
\(g: \mathbb{R} \rightarrow \mathbb{R}\) is a monotonically increasing function mapping preference scores \(s_w\) and \(s_l\) (for winning and losing responses \(y_w\) and \(y_l\)) into the logit space, \(\beta\) is a temperature parameter, \(\mathcal{D}\) represents the dataset distribution over input-output pairs \((x, y_w, y_l)\).

This formulation extends the DPO framework by incorporating preference differences \(g(s_w) - g(s_l)\), which reflect the utility of key regions identified by UV-CoT.

\subsection{DPO Background and Reparameterization}
DPO reformulates reward model training as a policy optimization problem by reparameterizing the reward function from Proximal Policy Optimization (PPO) \cite{schulman2017proximal}:
\begin{equation} \label{eq:dpo_reward}
r(x, y) = \beta \log \frac{\pi_{\theta}(y \mid x)}{\pi_{\mathrm{ref}}(y \mid x)} + \beta \log Z(x),
\end{equation}
where \(Z(x)\) is the partition function. Substituting this into the Bradley-Terry preference model \cite{david1963method} yields:
\begin{equation} \label{eq:bradley_terry}
\begin{aligned}
p(y_w \succ y_l) &= \sigma \left( r(x, y_w) - r(x, y_l) \right) \\
&= \sigma \left( \beta \log \frac{\pi_{\theta}(y_w \mid x)}{\pi_{\mathrm{ref}}(y_w \mid x)} - \beta \log \frac{\pi_{\theta}(y_l \mid x)}{\pi_{\mathrm{ref}}(y_l \mid x)} \right),
\end{aligned}
\end{equation}
where \(\sigma\) is the sigmoid function. Maximizing the log-likelihood of this preference model leads to the naive DPO loss:
\begin{equation} \label{eq:naive_dpo}
\begin{aligned}
\mathcal{L}_{\mathrm{DPO}}(\theta) &= -\mathbb{E}_{(x, y_w, y_l) \sim \mathcal{D}}  
\Bigg[ \log \sigma \Bigg( \beta \log \frac{\pi_{\theta}(y_w \mid x)}{\pi_{\mathrm{ref}}(y_w \mid x)}  \\
&\quad - \beta \log \frac{\pi_{\theta}(y_l \mid x)}{\pi_{\mathrm{ref}}(y_l \mid x)} \Bigg) \Bigg].
\end{aligned}
\end{equation}

\begin{table*}[t]
    \centering
    \caption{The details of the datasets, which spans five distinct domains and includes various source datasets.}
    \label{tab:visual_cot_dataset}
    \begin{tabular}{lccc}
        \toprule
        \textbf{Domain} & \textbf{Source Dataset} & \textbf{Size} &  \textbf{Dataset Description} \\
        \midrule
        \multirow{4}{*}{Text/Doc} & TextVQA & 16k & Images with text \\
        & DocVQA & 35k & Doc images \\
        & DUDE & 15k & Doc images \\
        & SROIE & 4k & Invoice images \\
        \midrule
        Chart & InfographicsVQA & 15k & Infographic images \\
        \midrule
        \multirow{2}{*}{General VQA} & Visual7W & 43k & Images \\
        & Flickr30k & 136k & Images\\
        \midrule
        \multirow{2}{*}{Relation Reasoning} & VSR & 3k & Images \\
        & GQA & 88k & Images \\
        \midrule
        High-Resolution & V$^*$ Bench & 238 & Images \\
        \bottomrule
    \end{tabular}
\end{table*}

\subsection{Derivation with Gumbel Distribution}
To incorporate preference-weighted optimization, we define \(\Delta_r = g(s_w) - g(s_l)\) and introduce Gumbel-distributed random variables \(R_w \sim \operatorname{Gumbel}(r(x, y_w), 1)\) and \(R_l \sim \operatorname{Gumbel}(r(x, y_l), 1)\). The probability that the winning response is preferred, adjusted by the preference gap, is:
\begin{equation} \label{eq:gumbel_prob}
\begin{aligned}
&P(R_w - R_l > \Delta_r) = \sigma \left( r(x, y_w) - r(x, y_l) - \Delta_r \right) \\
&= \sigma \left( \beta \log \frac{\pi^*(y_w \mid x)}{\pi_{\mathrm{ref}}(y_w \mid x)} - \beta \log \frac{\pi^*(y_l \mid x)}{\pi_{\mathrm{ref}}(y_l \mid x)} - \Delta_r \right).
\end{aligned}
\end{equation}
This result leverages the Gumbel-max trick \cite{Maddison2017Gumbel}, with a similar derivation found in ODTO \cite{amini2024direct}.

\subsection{Proof of Gumbel-Based Preference}
We first prove the foundational probability \(P(R_w - R_l > 0) = \sigma(\Delta_{\hat{r}_{\theta}})\), where \(\Delta_{\hat{r}_{\theta}} = r_{\theta}(x, y_w) - r_{\theta}(x, y_l)\).

\textbf{Proof:} Define the random variable \(I = \arg\max_{l, w} \{R_l, R_w\}\). The goal is to show:
\begin{equation} \label{eq:proof_target}
P(I = w) = \frac{\exp(\hat{r}_{\theta}(x, y_w))}{\exp(\hat{r}_{\theta}(x, y_w)) + \exp(\hat{r}_{\theta}(x, y_l))}.
\end{equation}
For notation simplicity, let \(\hat{r}_w = \hat{r}_{\theta}(x, y_w)\), \(\hat{r}_l = \hat{r}_{\theta}(x, y_l)\), and \(g_{\hat{r}_w} \sim \operatorname{Gumbel}(\hat{r}_w, 1)\). Then:
\begin{align}
&P(I = w) = \mathbb{E}_{m \sim g_{\hat{r}_w}} \left[ P(R_l < m) \right] \label{eq:proof_step1} \\
&= \int_{-\infty}^{\infty} \exp(-m) \cdot \exp\left(- \exp(\hat{r}_l - m) \right) \cdot \exp(\hat{r}_w) \, dm, \label{eq:proof_step2}
\end{align}
where the integral accounts for the Gumbel CDF. Let \(Z = \exp(\hat{r}_w) + \exp(\hat{r}_l)\). The expression simplifies to:
\begin{equation} \label{eq:proof_result}
P(I = w) = \frac{\exp(\hat{r}_w)}{Z} = \frac{\exp(\hat{r}_{\theta}(x, y_w))}{\exp(\hat{r}_{\theta}(x, y_w)) + \exp(\hat{r}_{\theta}(x, y_l))},
\end{equation}
proving Equation \eqref{eq:proof_target}. Extending this to the preference gap \(\Delta_r\), we derive:
\begin{equation} \label{eq:final_prob}
\begin{aligned}
P(R_w - R_l > \Delta_r) &= 1 - \mathcal{F}(\Delta_r) \\
&= \frac{1}{2} - \frac{1}{2} \tanh\left(\frac{\Delta_r - \Delta_{\hat{r}_{\theta}}}{2}\right) \\
&= \sigma(\Delta_{\hat{r}_{\theta}} - \Delta_r),
\end{aligned}
\end{equation}
where \(\Delta_{\hat{r}_{\theta}} = r_{\theta}(x, y_w) - r_{\theta}(x, y_l)\). This completes the derivation of the UV-CoT loss in Equation \eqref{eq:uvcot_loss}.

\section{Implement Details}
\subsection{Datasets}

To generate diverse and comprehensive image-level Chain-of-Thought (CoT) data for training Multimodal Large Language Models (MLLMs), we select nine source datasets spanning four distinct domains: Text/Doc, General Visual Question Answering (VQA), Charts, and Relation Reasoning. These domains are chosen to ensure a broad representation of visual reasoning tasks, enabling the model to develop robust CoT capabilities across varied contexts.

Before performing preference optimization, we conduct Supervised Fine-Tuning (SFT) using 10\% of the labeled Visual-CoT dataset, which corresponds to approximately 25k samples, as detailed in~\cref{tab:visual_cot_dataset}. This subset is chosen to balance computational efficiency with sufficient exposure to diverse reasoning patterns, resulting in a model we denote as \texttt{UV-CoT (10\%)}. Following SFT, we perform preference optimization using a total of 249k preference data points, curated from the same nine datasets. The preference data is generated by ranking model outputs for each dataset, ensuring that the distribution across domains mirrors that of Visual-CoT~\cite{shao2025visual}. Specifically, for each dataset, we maintain roughly the same proportion of preference pairs as in Visual-CoT (e.g., Text/Doc datasets contribute approximately 50\% of the data, consistent with their representation in our dataset). This approach ensures that \texttt{UV-CoT} benefits from a balanced and comprehensive preference optimization process, enhancing its ability to prioritize key regions in visual reasoning tasks.

As shown in \cref{tab:visual_cot_dataset}, we provide a detailed introduction to the datasets we used. For the Text/Doc domain, we include four datasets focusing on text recognition and comprehension in diverse document and image formats: DocVQA~\cite{mathew2021docvqa}, TextVQA~\cite{singh2019towards}, DUDE~\cite{van2023document}, and SROIE~\cite{huang2019icdar2019}. These datasets provide rich text-based reasoning scenarios, such as extracting information from invoices (SROIE) or answering questions about document content (DocVQA), which are crucial for generating CoT data that involves step-by-step text interpretation.

In the General VQA domain, we select Flickr30k~\cite{plummer2015flickr30k} and Visual7W~\cite{zhu2016visual7w}. These datasets are well-suited for general visual question answering, as they contain diverse images paired with questions that require understanding both visual content and textual prompts, facilitating the creation of CoT data for general reasoning tasks.

For the Charts domain, we use the InfographicsVQA dataset~\cite{mathew2022infographicvqa}, which consists of high-resolution infographics. This dataset is particularly advantageous for training MLLMs to localize and interpret specific regions in charts, enabling the generation of CoT data that involves reasoning about data visualization elements such as legends, labels, and trends.

In the Relation Reasoning domain, we select VSR~\cite{liu2023visual} and GQA~\cite{hudson2019gqa}. These datasets are rich in spatial relational information among objects in images, making them ideal for constructing CoT data that focus on reasoning about object relationships, such as identifying relative positions or dependencies in a scene.

For high-resolution image reasoning, we use V$^*$ Bench \cite{wu2024v}, which comprises 238 images from the SA-1B dataset~\cite{kirillov2023segment} with an average resolution of 2246$\times$1582.

\subsection{Evaluation}


We utilize GPT-4o to assess the performance of our model due to its superior reasoning capabilities and adopt an evaluation prompt to. The prompt template is like:
\begin{tcolorbox}[
    colback=gray!20, 
    colframe=black!70, 
    arc=2mm, 
    auto outer arc, 
    boxrule=0.3mm, 
    width=0.47\textwidth, 
    title=\centering Evaluating Prompt Template
]
\textit{
You are a helpful and precise assistant for checking the quality of the answer, you need to give a score to the model's answer by referring to the standard answer, based on the given question. The full score is 1 point and the minimum score is 0 points. Please output the score in the form `score: \textless score \textgreater'. The evaluation criteria require that the closer the model's answer is to the standard answer, the higher the score.}
\end{tcolorbox}

The meaning score, ranging from 0 to 1, reflects the semantic relevance of the model's responses to the given prompts.



\section{Limitations}
While our UV-CoT model demonstrates high performance across most evaluated datasets, it encounters challenges in accurately identifying anchor boxes on certain datasets, notably DocVQA~\cite{mathew2021docvqa} and InfographicsVQA~\cite{mathew2022infographicvqa}. These difficulties may arise due to the complex layouts, variable text densities, and noisy annotations prevalent in these datasets, which complicate the localization of relevant regions. In contrast, the ground truth (GT) boxes achieve exceptional performance on these datasets, suggesting that our model has significant untapped potential. Future research could explore advanced anchor box detection algorithms, such as incorporating adaptive thresholding or multi-scale feature extraction, to address these limitations and enhance the model's robustness across diverse visual domains.

\section{Potential negative societal impacts}

The potential negative societal impacts of our work align with those of other MLLMs and LLMs. While the development of UV-CoT and MLLMs advances AI capabilities, it also introduces several risks. These include heightened privacy concerns, the reinforcement of existing biases, the spread of misinformation, job displacement due to automation, and ethical challenges related to accountability, transparency, and informed consent. Addressing these issues requires responsible deployment, continuous monitoring, and the implementation of safeguards to mitigate unintended consequences.

\end{document}